\documentclass[11pt,a4paper]{article}
\usepackage[hyperref]{naaclhlt2018}
\usepackage{times}
\usepackage{latexsym}
\usepackage{graphicx}
\usepackage{url}
\usepackage{amsmath}

\usepackage{latexsym}
\usepackage{graphicx}
\usepackage{url}
\usepackage{amsmath}
\usepackage{float}
\usepackage{adjustbox}
\usepackage[shortlabels]{enumitem}
\usepackage{subcaption}
\usepackage{hyperref}
\usepackage{amsfonts}
\aclfinalcopy 

\DeclareMathOperator*{\trr}{{\color{blue} \triangleright}}

\newcommand{\ovr}{\overrightarrow}
\newcommand{\ovl}{\overleftarrow}

\newcommand{\embed}{\text{embed}}
\newcommand{\cont}{\text{c}}

\newcommand{\mg}{\text{sg}}

\DeclareGraphicsExtensions{.pdf,.png,.jpg}
\title{Dynamic Fusion Networks for Machine Reading Comprehension
}

\author{Yichong Xu$^1$\thanks{ \;\;Most of this work was performed when the author was interning at Microsoft AI\&R.}, Jingjing Liu$^{2}$, Jianfeng Gao$^{2}$, Yelong Shen$^{2}$ and Xiaodong Liu$^{2}$ \\
	$^1$ Carnegie Mellon University\\
	$^2$ Microsoft AI\&R\\
	\texttt{yichongx@cs.cmu.edu}\\
	\texttt{\{jingjl, jfgao, yeshen, xiaodl\}@microsoft.com } }

\date{}
\begin{document}
\maketitle
\begin{abstract}
This paper presents a novel neural model - Dynamic Fusion Network (DFN), for machine reading comprehension (MRC). DFNs differ from most state-of-the-art models in their use of a dynamic multi-strategy attention process, in which passages, questions and answer candidates are jointly fused into attention vectors, along with a dynamic multi-step reasoning module for generating answers. With the use of reinforcement learning, for each input sample that consists of a question, a passage and a list of candidate answers, an instance of DFN with a sample-specific network architecture can be dynamically constructed by determining what attention strategy to apply and how many reasoning steps to take. Experiments show that DFNs achieve the best result reported on RACE, a challenging MRC dataset that contains real human reading questions in a wide variety of types. A detailed empirical analysis also demonstrates that DFNs can produce attention vectors that summarize information from questions, passages and answer candidates more effectively than other popular MRC models. 
\end{abstract}

\section{Introduction}

The goal of Machine Reading Comprehension (MRC) is to have machines read a text passage and then generate an answer (or select an answer from a list of given candidates) for any question about the passage. There has been a growing interest in the research community in exploring neural MRC models in an end-to-end fashion, thanks to the availability of large-scale datasets, such as CNN/DM \cite{hermann2015teaching} and SQuAD \cite{rajpurkar2016squad}. 

Despite the variation in model structures, most state-of-the-art models perform reading comprehension in two stages. First, the symbolic representations of passages and questions are mapped into vectors in a neural space. This is commonly achieved via embedding and attention \cite{seo2016bidirectional,weissenborn2017making} or fusion \cite{huang2017fusionnet}. Then, reasoning is performed on the vectors to generate the right answer.

Ideally, the best attention and reasoning strategies should adapt organically in order to answer different questions.
However, most MRC models use a static attention and reasoning strategy indiscriminately, regardless of various question types. One hypothesis is because these models are optimized on those datasets whose passages and questions are domain-specific (or of a single type). For example, in CNN/DM, all the passages are news articles, and the answer to each question is an entity in the passage. In SQuAD, the passages came from Wikipedia articles and the answer to each question is a text span in the article. Such a fixed-strategy MRC model does not adapt well to other datasets.
For example, the exact-match score of BiDAF \cite{seo2016bidirectional}, one of the best models on SQuAD, drops from 81.5 to 55.8 when applied to TriviaQA \cite{joshi2017triviaqa}, whereas human performance is 82.3 and 79.7 on SQuAD and TriviaQA, respectively.



In real-world MRC tasks, we must deal with questions and passages of different types and complexities, which calls for models that can dynamically determine what attention and reasoning strategy to use for any input question-passage pair on the fly. In a recent paper, \cite{shen2017reasonet} proposed dynamic multi-step reasoning, where the number of reasoning steps is determined spontaneously (using reinforcement learning) based on the complexity of the input question and passage. 
With a similar intuition, in this paper
we propose a novel MRC model which is dynamic not only 
on the \emph{number} of reasoning steps it takes, but also on the \emph{way} it performs attention.
To the best of our knowledge, this is the first MRC model with this dual-dynamic capability.

The proposed model is called a Dynamic Fusion Network (DFN). In this paper, we describe the version of DFN developed on the RACE dataset \cite{lai2017race}. In RACE, a list of candidate answers is provided for each passage-question pair. So DFN for RACE is a scoring model - the answer candidate with the highest score will be selected as the final answer. 

Like other MRC models, DFNs also perform machine reading in two stages: attention and reasoning. DFN is unique in its use of a dynamic multi-strategy attention process in the attention stage. 
Here ``attention'' refers to the process that texts from different sources (passage, question, answers) are combined in the network. 
In literature, a fixed attention mechanism is usually employed in MRC models. In DFN, the attention strategy 
is not static; instead, the actual strategy for drawing attention among the three text sources 
are chosen \emph{on the fly} for each sample. This lends flexibility to adapt to various question types that require different comprehension skills. The output of the attention stage is then fed into the reasoning module to generate the answer score. The reasoning module in DFN uses dynamic multi-step reasoning, where the number of steps depends on the complexity of the question-passage pair and varies from sample to sample. 

Inspired by ReasoNet \cite{shen2017reasonet} and dynamic neural module networks \cite{andreas2016learning}, we use deep reinforcement learning methods \cite{mnih2013playing,zhang2017learning} to dynamically choose the optimal attention strategy and the optimal number of reasoning steps for a given sample. We use RL in favor of other simpler methods (like cascading, pooling or weighted averaging) mainly because we intend to learn a policy that constructs an instance of DFN of a sample-specific structure. Given an input sample consisting of a question, a passage and a list of candidate answers in RACE, an instance of DFN can be constructed via RL step by step on the fly. Such a policy is particularly appealing as it also provides insights on how the model performs on different types of questions. At each decision step, the policy maps its ``state'', which represents an input  sample, and DFN's \emph{partial} knowledge of the right answer, to the action of assembling proper attention and reasoning modules for DFN.

Experiments conducted on the RACE dataset show that DFN significantly outperforms previous state-of-the-art MRC models and has achieved the best result reported on RACE. A thorough empirical analysis also demonstrates that DFN is highly effective in understanding passages of a wide variety of styles and answering questions of different complexities.

\section{\label{sec:relatedwork}Related Work}

The recent progress in MRC is largely due to the introduction of large-scale datasets. CNN/Daily Mail \cite{hermann2015teaching} and SQuAD \cite{rajpurkar2016squad} are two popular and widely-used datasets. More recently, other datasets using different collection methodologies have been introduced, such as MS MARCO \cite{nguyen2016ms}, NewsQA \cite{trischler2016newsqa} and RACE \cite{lai2017race}. For example, MS MARCO collects data from search engine queries and user-clicked results, thus contains a broader topic coverage than Wikipedia and news articles in SQuAD and CNN/Daily Mail. 
Among the large number of MRC datasets, RACE focuses primarily on developing MRC models with near-human capability. Questions in RACE come from real English exams designed specifically to test human comprehension. This makes RACE an appealing testbed for DFN; we will further illustrate this in Section \ref{sec:analysis}.

The word ``fusion'' for MRC was first used by FusionNet \cite{huang2017fusionnet} to refer to the process of updating the representation of passage (or question) using information from the question (or passage) representation. A typical way of fusion is through attention: for example, BiDAF \cite{seo2016bidirectional} uses a bi-directional attention, where the representation of passage (or question) vectors are re-weighted by their similarities to the question (or passage) vectors. We will use ``fusion'' and ``attention'' interchangeably throughout the paper.

In the attention process of state-of-the-art MRC models, a pre-defined attention strategy is often applied. \cite{wang2017bilateral} proposed a Bi-directional Multi-Perspective Matching (BiMPM) model, which uses attention with multiple perspectives characterized by different parameters. Although multi-perspective attention might be able to handle different types of questions, all perspectives are used for all the questions. DFN is inspired by BiMPM, but our dynamic attention process is more adaptive to variations of questions. 



Another important component of MRC systems is the answer module, which performs reasoning to generate the final prediction. The reasoning methods in existing literature can be grouped into three categories:
1) single-step reasoning \cite{chen2016thorough,chen2017reading,seo2016bidirectional,wang2017gated}; 2) multi-step reasoning with a fixed number of steps \cite{dhingra2016gated,sordoni2016iterative,xiong2016dynamic}; and 3)  \emph{dynamic} multi-step reasoning (ReasoNet \cite{shen2017reasonet}). In particular, \cite{xiong2016dynamic} proposed handling the variations in passages and questions using Maxout units and iterative reasoning. However, this model still applies static attention and reasoning (with fixed multiple steps), where the same attention strategy is applied to all questions.
DFN can be seen as an extension of ReasoNet, in the sense that the dynamic strategy is applied not only in the reasoning process but also in the attention process.

The idea of dynamic attention has been applied to article recommendations \cite{wang2017dynamic}. For MRC, Andreas et al. (2016) proposed a dynamic decision process for reading comprehension task~\cite{andreas2016learning}. In their dynamic neural module networks, the MRC task is divided into several predefined steps (e.g., finding, lookup, relating), and a neural network is dynamically composed  via RL based on parsing information. In DFN, we also incorporate dynamic decisions, but instead of using fixed steps, we apply dynamic decisions to various attention strategies and flexible reasoning steps.

\section{\label{sec:analysis} RACE - The MRC Task}

In this section, we first give a brief introduction to the RACE dataset,
and then explain the rationale behind choosing RACE as the testbed in our study.

\subsection{\label{sec:briefIntroRace} The Dataset}
RACE (Reading Comprehension Dataset From Examinations) is a recently released MRC dataset consisting of 27,933 passages and 97,867 questions from English exams, targeting Chinese students aged 12-18. RACE consists of two subsets, RACE-M and RACE-H, from middle school and high school exams, respectively. 
RACE-M has 28,293 questions and RACE-H has 69,574. 
Each question is associated with 4 candidate answers, one of which is correct. The data generation process of RACE differs from most MRC datasets - instead of generating questions and answers by heuristics or crowd-sourcing, questions in RACE are specifically designed for testing human reading skills, and are created by domain experts. 

\subsection{\label{sec:mainadv}Distinctive Characteristics in RACE}

The RACE dataset has some distinctive characteristics compared to other datasets, making it an ideal testbed for developing generic MRC systems for real-world human reading tasks.


\begin{figure}
	\begin{center}
		\includegraphics[angle=0]{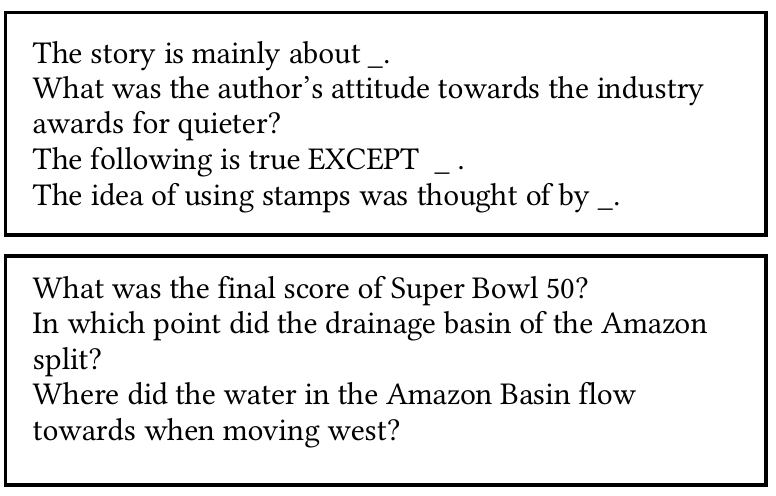}	
	\end{center}
	
	\caption{\label{fig:compques}Above: Questions from RACE. Below: Questions from SQuAD. 
	}
\end{figure}

\begin{figure*}
	\begin{center}
		\includegraphics[width=1.0\textwidth]{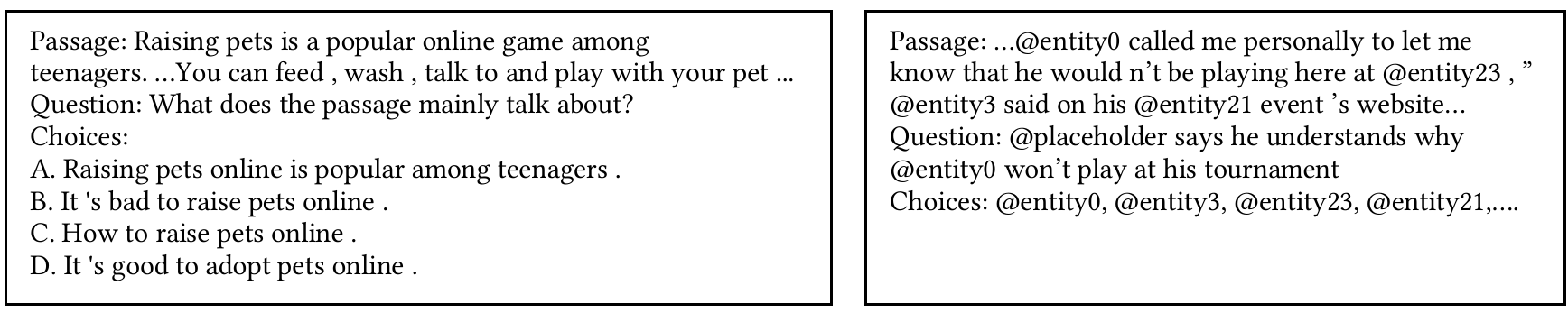}	
	\end{center}
	
	\caption{\label{fig:compans}Left: Examples from RACE dataset. Right: Examples from CNN dataset.}

\end{figure*}

\textbf{Variety in Comprehension Skills.} 
RACE requires a much broader spectrum of comprehension skills than other MRC datasets. Figure \ref{fig:compques} shows some example questions from RACE and SQuAD: most SQuAD questions lead to direct answers that can be found in the original passage, while questions in RACE require more sophisticated reading comprehension skills such as summarizing (1st question), inference (2nd question) and deduction (3rd question). For humans, various tactics and skills are required to answer different questions.
Similarly, it is important for MRC systems to adapt to different question types.


\textbf{Complexity of Answers.} As shown in Figure \ref{fig:compans}, the answers in CNN/DM dataset are entities only. In SQuAD-like datasets, answers are often constrained to spans in the passage. Different from these datasets, answer candidates in RACE are natural language sentences generated by human experts, which increases the difficulty of the task. Real-world machine reading tasks are less about span exact matching, and more about summarizing the content and extending the obtained knowledge through reasoning. 

\textbf{Multi-step reasoning.} Reasoning is an important skill in human reading comprehension. It refers to the skill of making connection between sentences and summarizing information throughout the passage. Table \ref{tab:inference} shows a comparison on the requirement of reasoning level among different datasets. The low numbers on SQuAD and CNN/DM show that reasoning skills are less critical in getting the correct answers in these datasets, whereas such skills are essential for answering RACE questions.

\begin{table*}
	\begin{center}
		\begin{tabular}{|@{\hskip2pt} c @{\hskip2pt}|@{\hskip2pt} c @{\hskip2pt}|@{\hskip2pt} c @{\hskip2pt}|}
			\hline \bf Dataset & Single-sentence Reasoning &  Multi-sentence Reasoning  \\ \hline
			CNN & 19.0\% & 2.0\%\\
			SQuAD &8.6\% & 11.9\%\\
			NewsQA & 13.2\% & 20.7\%\\
			RACE-M & 31.3\% &22.6\%\\
			RACE-H & 34.1\% & 26.9\%\\
			RACE & 33.4\% & 25.8\%\\
			\hline
		\end{tabular}
	\end{center}
	\caption{\label{tab:inference} Percentage of questions in each dataset that require Single-sentence Reasoning and Multi-sentence Reasoning \cite{lai2017race}.}
	
\end{table*}
%

\section{\label{sec:model}Dynamic Fusion Networks}

In this section, we present the model details of DFN. Section \ref{sec:modelarch} describes the overall architecture, and each component is explained in detail in subsequent subsections. Section \ref{sec:trainDetail} describes the reinforcement learning methods used to train DFN.

\subsection{\label{sec:modelarch}Model Architecture}

\begin{figure*}[ht]
	\begin{center}
		\includegraphics[width=0.98\textwidth]{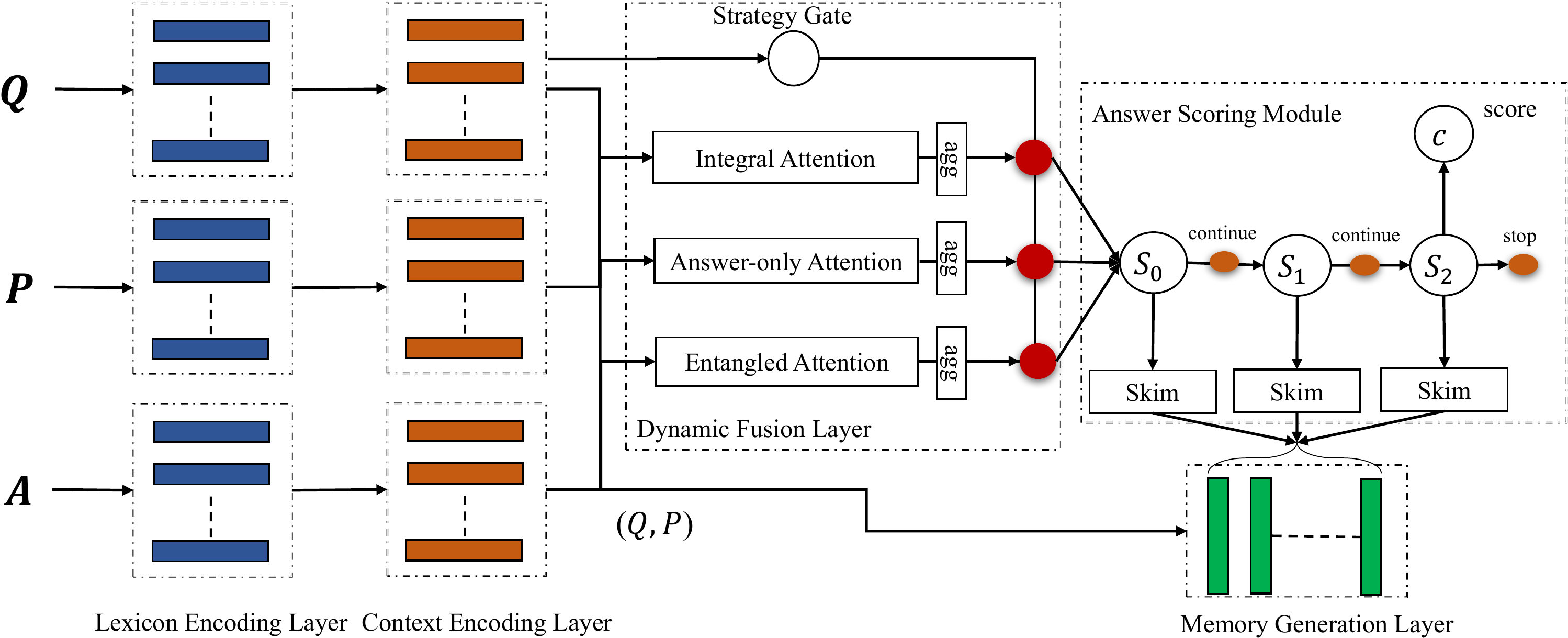}	
	\end{center}
	
	\caption{\label{fig:musicmodel}Architecture of DFN. For simplicity, we only draw DFN for one answer candidate $A$. 
		i) Passage, question and answer candidates are independently mapped through word and character encodings in the Lexicon Encoding Layer. ii) The independent encodings are then fed into a BiLSTM in the Context Encoding Layer. iii) The Dynamic Fusion Layer takes a customized attention strategy across the three representations of passage, question and answer candidates. iv) Memory Generation Layer generates a working memory. v) The Answer Scoring Module reads in the memory for a dynamic number of steps. vi) Answer prediction module generates the final output.
	}

\end{figure*}

The overall architecture of DFN is depicted by Figure \ref{fig:musicmodel}. The input is a question $Q$ in length $l_q$, a passage $P$ in length $l_p$, and a list of $r$ answer candidates  $\mathcal{A}=\{A_1,...,A_r\}$ in length $l_{a}^1,...,l_a^r$. 
The model produces scores $c_1, c_2, ..., c_r$ for each answer candidate $A_1, A_2, ..., A_r$ respectively. 
The final prediction module selects the answer with the highest score.

The architecture consists of a standard Lexicon Encoding Layer and a Context Encoding Layer, 
on top of which are a Dynamic Fusion Layer and a Memory Generation Layer. The Dynamic Fusion Layer applies different attention strategies to different question types, and the Memory Generation Layer encodes question-related information in the passage for answer prediction.
Multi-step reasoning is conducted over the output from the Dynamic Fusion and Memory Generation layers, in the Answer Scoring Module. The final output of the model is an answer choice $C\in \{1,2,...,r\}$ from the Answer Prediction Module. 

In the following subsections, we will describe the details of each component in DFN (bold letters represent trainable parameters).

\subsection{Lexicon Encoding Layer}
The first layer of DFN transforms each word in the passage, question and answer candidates independently into a fixed-dimension vector. This vector is the concatenation of two parts. The first part is the pre-trained GloVe embedding \cite{pennington2014glove} of each word. For each out-of-vocabulary word, we map it to an all-zero vector. The second part is the character encodings. This is carried out by mapping each character to a trainable embedding, and then feeding all characters into an LSTM \cite{hochreiter1997long}. The last state of this LSTM is used as the character encodings. The output of the Lexicon Encoding layer is a set of vectors for $Q,P$ and each answer candidate in $\mathcal{A}$, respectively: $Q^\embed=\{q^\embed_i\}_{i=1}^{l_q}, P^\embed=\{p^\embed_i\}_{i=1}^{l_p}$, and $A^\embed_j=\{a^\embed_{i,j}\}_{i=1}^{l_a^j}, j=1,2,...,r$.

\subsection{Context Encoding Layer}
The Context Encoding Layer passes $Q^{\embed}, p^{\embed}$ and $A^{\embed}$ into a bi-directional LSTM (BiLSTM) to obtain context representations. Since answer candidates $A_1,...,A_r$ are not always complete sentences, we append the question before each answer candidate and feed the concatenated sentence into BiLSTM. We use the same BiLSTM to encode the information in $P,Q$ and $\mathcal{A}$. The obtained context vectors are represented as:
\begin{align*}
	&Q^\cont=\textbf{BiL}\textbf{STM}_1(Q^\embed)=\{\overrightarrow{q_i^\cont},\overleftarrow{q_i^\cont}\}_{i=1}^{l_q},\\
	&P^\cont=\textbf{BiL}\textbf{STM}_1(P^\embed)=\{\overrightarrow{p_i^\cont},\overleftarrow{p_i^\cont}\}_{i=1}^{l_p},\\
	&(Q+A)^\cont_j=\textbf{BiLSTM}_1(Q^\embed+A^\embed_j)\\
	&=\{\overrightarrow{a^\cont_{i,j}},\overleftarrow{a^\cont_{i,j}}\}_{i=1}^{l_p+l_a^j}, j=1,2,...,r.
\end{align*}

\subsection{Dynamic Fusion Layer} 
This layer is the core of DFN. For each given question-passage pair, one of $n$ different attention strategies is selected to perform attention across the passage, question and answer candidates.

The dynamic fusion is conducted in two steps: in the first step, an attention strategy $G\in \{1,2,...,n\}$ is randomly sampled from the output of the \emph{strategy gate} $f^\mg(Q^c)$ . The strategy gate takes input from the last-word representation of the question $Q^\cont$, and outputs a softmax over $\{1,2,...n\}$. In the second step, the $G$-th attention strategy is activated, and computes the attention results according to $G$-th strategy. Each strategy, denoted by \textbf{Attention}$_k,k=1,2,...,n$, is essentially a function of $Q^\cont,P^\cont$ and one answer candidate $(Q+A)^\cont_j$ that performs attention in different directions.
The output of each strategy is 
a fixed-dimension representation, as the attention result\footnote{This model can easily be extended, to have each strategy produce a variable-length vector.}.

\begin{align*}
	f^\mg(Q^c) \leftarrow & \text{softmax}(\mathbf{W_1}(\overrightarrow{q_{l_q}^\cont};\overleftarrow{q_{1}^\cont})\\
	G\sim & \text{Category}\left(f^\mg(Q^c)\right),\\
	s_j\leftarrow & \textbf{Attention}_{G}(Q^\cont,P^\cont,(Q+A)^\cont_j),\\
	& j=1,2,...,r.
\end{align*}
\begin{figure*}[ht]
	\centering
	\begin{subfigure}[b]{0.2 \textwidth} 
		\centering
		\includegraphics[height=1.6cm]{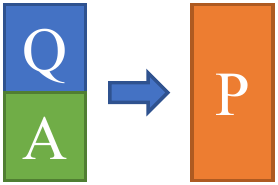}
		\caption{Integral Attention}
		\label{fig:intatt}
	\end{subfigure}
	\begin{subfigure}[b]{0.27 \textwidth} 
		\centering
		\includegraphics[height=3.5cm]{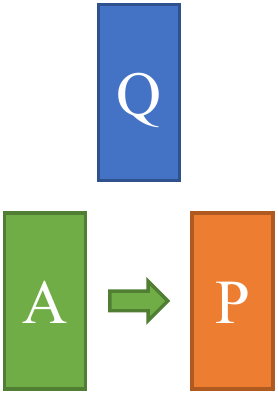}
		\caption{Answer-only Attention}
		\label{fig:asoatt}
	\end{subfigure}
	\begin{subfigure}[b]{0.27\textwidth} 
		\centering
		\includegraphics[height=3.5cm]{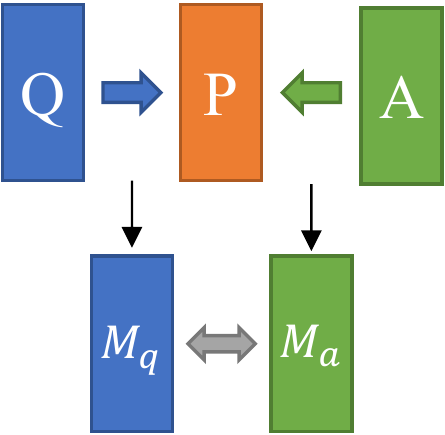}
		\caption{Entangled Attention}
		\label{fig:entatt}
	\end{subfigure}
	\begin{subfigure}[b]{0.2 \textwidth} 
		\centering
		\includegraphics[height=1.6cm]{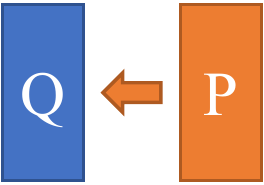}
		\caption{Memory Generation}
		\label{fig:mematt}
	\end{subfigure}

	\caption[Graphical description]{Diagram illustrating the Attention Strategies and Memory Generation in DFN.}
	\label{fig:attfig}

\end{figure*}

\noindent\textbf{Attention Strategies.}
For experiment on RACE, we choose $n=3$ and use the following strategies:
\begin{enumerate}[label={(\arabic*)},wide=0pt]
	\item \textbf{Integral Attention}: We treat the question and answer as a whole, and attend each word in $(Q+A)_j^\cont$ to the passage $P^\cont$ (Figure \ref{fig:intatt}). 
	This handles questions with short answers (e.g., the last question in upper box of Figure \ref{fig:compques}). 
	
	Formally,
	
	\[Q^{\text{int}}_{j},A^{\text{int}}_{j}\leftarrow\text{Split}\left((Q+A)^\cont_j\trr P^\cont \right).
	\]
	The operator $\trr$ represents any one-sided attention function. For DFN, we use the single direction version of multi-perspective matching in BiMPM \cite{wang2017bilateral}; For two text segments $X,X'\in \{P,Q,A_j,(Q+A)_j\}$, $X \trr X'$ matches each word $w\in X$ with respect to the whole sentence $X'$, and has the same length as $X$. 
	We defer details of the $\trr$ operator to Section \ref{sec:memgen} when we introduce our memory generation.
	
	The Split$()$ function splits a vector representation in length $l_q+l_a^j$ into two vector representations in length $l_q$ and $l_a^j$, to be consistent with other strategies.
	\item \textbf{Answer-only Attention}: This strategy only attends each word in the answer candidate to the passage (Figure \ref{fig:asoatt}), without taking the question into consideration. This is to handle questions with full-sentence answer candidates (e.g., the first and the third questions in the upper box of Figure \ref{fig:compques}). 
	\begin{align*}	
		M_a\leftarrow& A_j^\cont \trr P^\cont,\\
		Q_j^{\text{aso}},A_j^{\text{aso}}\leftarrow&Q^\cont,M_a.
	\end{align*}
	\item \textbf{Entangled Attention}: As shown in Figure \ref{fig:entatt}, each word in question and answer is attended to the passage, denoted by $M_q$ and $M_a$. 
	Then, we entangle the results by attending each word in $M_q$ to $M_a$, and also $M_a$ to $M_q$. This attention is more complicated than the other two mentioned above, and targets questions that require reasoning (e.g., the second question in the upper box of Figure \ref{fig:compques}).
	\begin{align*}
		M_q\leftarrow& Q^\cont \trr P^\cont\\
		M_a\leftarrow& A_j^\cont \trr P^\cont,\\
		Q_j^{\text{ent}},A_j^{\text{ent}}\leftarrow&M_q \trr M_a, M_a \trr M_q.
	\end{align*}
\end{enumerate}
We can incorporate a large number of strategies into the framework depending on the question types we need to deal with. In this paper, we use three example strategies to demonstrate the effectiveness of DFN.

\noindent\textbf{Attention Aggregation.} Following previous work, we aggregate the result of each attention strategy through a BiLSTM. The first and the last states of these BiLSTMs are used as the output of the attention strategies. We use different BiLSTM for different strategies, which proved to slightly improve the model performance.
\begin{align*}
	Q_j^x,A_j^x\leftarrow&\textbf{BiLSTM}^x(Q_j^x), \textbf{BiLSTM}^x(A_j^x),\\
	\textbf{Attention}_k\leftarrow& \text{FinalState}(Q_j^x,A_j^x),\\
	&\text{   for } (k,x)\in \{ \text{(1,int),(2,aso),(3,ent)}\}.
\end{align*}

The main advantages of dynamic multi-strategy fusion are three-fold: 1) It provides adaptivity for different types of questions. This addresses the challenge in the rich variety of comprehension skills aforementioned in Section \ref{sec:mainadv}. 
The key to adaptivity is the strategy gate $G$. Our observation is that the model performance degrades when trained using simpler methods such as max-pooling or model averaging. 2) The dynamic fusion takes all three elements (question, passage and answer candidates) into account in the attention process. This way, answer candidates are fused together with the question and the passage to get a complete understanding of the full context. 
3) There is no restriction on the attention strategy used in this layer, which allows flexibility for incorporating existing attention mechanisms.


Although some of the attention strategies appear to be straightforward (e.g., long/short answers), it is difficult to use simple heuristic rules for strategy selection. For example, questions with a placeholder ``\_'' might be incomplete question sentences that require integral attention; but in some questions (e.g., ``we can infer from the passage that \_ .''), the choices are full sentences and the answer-only attention should be applied here instead. Therefore, we turn to reinforcement learning methods (see Section \ref{sec:trainDetail}) to optimize the choice of attention strategies, which leads to a policy that give important insights on our model behavior.

\subsection{\label{sec:memgen}Memory Generation Layer} 
A memory is generated for the answer module in this layer. The memory $M$ has the same length as $P$, and is the result of attending each word in $P^\cont$ to the question $Q^\cont$ (Figure \ref{fig:mematt}). We use the same attention function for $M$ as that for attention strategies, and then aggregate the results. The memory is computed as $M\leftarrow\textbf{BiLSTM}_2(Q^\cont \trr P^\cont)$, where $\trr$ is the attention operator specified as below.

Our attention operator takes the same form as BiMPM \cite{wang2017bilateral}. For simplicity, we use $P,Q,(Q+A)_j$ to denote $P^\cont,Q^\cont$ and $(Q+A)^\cont_j$ in this section. Recall that for $X,X'\in \{P,Q,A_j,(Q+A)_j\}$, and $X \trr X'$ computes the relevance of each word $w\in X$ with respect to the whole sentence $X'$. $X \trr X'$ has the same length as $X'$. Each operation $\trr$ is associated with a set of trainable weights denoted by $\mathbf{W}_{o1},...,\mathbf{W}_{o8}$. For $\trr$ in different strategies, we use different sets of trainable weights; the only exception is for $M_a$ computed both in Answer-only Attention and Entangled Attention: These two operations have the same weights since they are exactly the same. We find untying weights in different $\trr$ operations can slightly improve our model performance. 

We use a multi-perspective function to describe $\trr$. For any two vectors $v_1,v_2\in \mathbb{R}^d$, define the multi-perspective function 
\[g(v_1,v_2;\mathbf{W})=\left\{\cos(\mathbf{W}^{(k)} \circ v_1, \mathbf{W}^{(k)}\circ v_2 )\right\}_{k=1}^N, \]
where $\mathbf{W}\in \mathbb{R}^{N\times d}$ is a trainable parameter, $N$ is a hyper-parameter (the number of perspectives), and $\mathbf{W}^{(k)}$ denotes the $k$-th row of $\mathbf{W}$. In our experiments, we set $N=10$. 

Now we define $X \trr X'$ using $g$ and four different ways to combine vectors in text $X,X'$. Denote by $x_i,x_i'\in \mathbb{R}^d$ the $i$-th vector in $X,X'$ respectively. The function work concurrently for the forward and backward LSTM activations (generated by BiLSTM in the Context Encoding layer) in $X$ and $X'$; denoted by $\overrightarrow{x}_i$ and $\overleftarrow{x}_i$, the forward and backward activations respectively (and similarly for $x_i'$). The output of $\trr$ also has activations in two directions for further attention operation (e.g., in Entangled Attention). The two directions are concatenated before feeding into the aggregation BiLSTM. 

Let $l_x,l'_x$ be the length of $X,X'$ respectively. $X\trr X'$ outputs two groups of vectors $\{\overrightarrow{u}_i,\overleftarrow{u}_i \}_{i=1}^{l_x}$ by concatenating the following four parts below:
\begin{enumerate}[i)]
	\item Full Matching: 
	\[\ovr{u}_i^{\text{full}}=g(\ovr{x}_i,\ovr{x}_{l_x'},\mathbf{W}_{o1}),\]
	\[\ovl{u}_i^{\text{full}}=g(\ovl{x}_i,\ovl{x}'_{1},\mathbf{W}_{o2}). \]
	\item Maxpooling Matching:
	\[\ovr{u}_i^{\text{max}}=\max_{j\in \{1,...,l_x\}}g(\ovr{x}_i,\ovr{x}_j',\mathbf{W}_{o3}),\]
	\[\ovl{u}_i^{\text{max}}=\max_{j\in \{1,...,l_x\}}g(\ovl{x}_i,\ovl{x}_j',\mathbf{W}_{o4}),\]
	here $\max$ means element-wise maximum.
	\item Attentive Matching: for $j=1,2,...,N$ compute
	\[\ovr{\alpha}_{i,j}=\cos(\ovr{x}_i,\ovr{x}_j'),\ovl{\alpha}_{i,j}=\cos(\ovl{x}_i,\ovl{x}_j'). \]
	Take weighted mean according to $\ovr{\alpha}_{i,j},\ovl{\alpha}_{i,j}$:
	\[\ovr{x}_i^{\text{mean}}=\frac{\sum_{j=1}^{l_x'} \ovr{\alpha}_{i,j} \cdot \ovr{x}_j'}{\sum_{j=1}^{l_x'} \ovr{\alpha}_{i,j}},\]
	\[\ovl{x}_i^{\text{mean}}=\frac{\sum_{j=1}^{l_x'} \ovl{\alpha}_{i,j} \cdot \ovl{x}_j'}{\sum_{j=1}^{l_x'} \ovl{\alpha}_{i,j}}.\]
	Use multi-perspective function to obtain attentive matching:
	\[\ovr{u}_i^{\text{att}}=g(\ovr{x}_i,\ovr{x}_i^{\text{mean}},\mathbf{W}_{o5}),\]
	\[\ovl{u}_i^{\text{att}}=g(\ovl{x}_i,\ovl{x}_i^{\text{mean}},\mathbf{W}_{o6}).\]
	\item Max-Attentive Matching: The same as attentive matching, but taking the maximum over $\ovr{\alpha}_{i,j},\ovl{\alpha}_{i,j}, j=1,2,...,l_x'$ instead of using the weighted mean.
\end{enumerate}

\subsection{Answer Scoring Module} 
This module performs multi-step reasoning in the neural space to generate the right answer.
This unit adopts the architecture of ReasoNet \cite{shen2017reasonet}. 
We simulate multi-step reasoning with a GRU cell \cite{cho2014properties} to skim through the memory several times, changing its internal state as the skimming progresses. The initial state $s_j^{(0)}=s_j$ is generated from the Dynamic Fusion Layer for each answer candidate $j=1,2,...,r$. We skim through the passage for at most $\mathcal{T}_{\max}$ times. In every step $t\in \{1,2,...,\mathcal{T}_{\max}\}$, an attention vector $f^{(t)}_{\text{att}}$ is generated from the previous state $s_j^{t-1}$ and the memory $M$. To compute $f_{\text{att}}$, an attention score $a_{t,i}$ is computed based on each word $m_i$ in memory $M$ and state $s^{(t)}_j$ as 

$$a_{i,j}^{(t)}\leftarrow \text{softmax}_{i=1,2,...,l_m} \lambda\cos\left(\mathbf{W_2}m_i, \mathbf{W_3}s^{(t)}_j\right)$$
where $l_m=l_p$ is the memory length, and $\mathbf{W_2},\mathbf{W_3}$ are trainable weights. We set $\lambda=10$ in our experiments. The attention vector is then computed as a weighted sum of memory vectors using attention scores, i.e., $f^{(t)}_{\text{att}}\leftarrow \sum_{i=1}^{l_m} a_{i,j}^{(t)}m_i.$
Then, the GRU cell takes the attention vector $f_{\text{att}}^{(t)}$ as input and changes its internal state.

\begin{align*}
	s_j^{(0)}\leftarrow s_j, \;s_j^{(t)}\leftarrow\textbf{GRU}\left(f_{\text{att}}^{(t)},s_j^{(t-1)}\right).
\end{align*}
To decide when to stop skimming, a termination gate (specified below) takes $s_j^{(t)}, j=1,...,r$ at step $t$ as the input, and outputs a probability $p_t$ of whether to stop reading.
The number of reading steps is decided by sampling a Bernoulli variable $T_t$ with parameter $p_t$. If $T_t$ is 1, the Answer Scoring Module stops skimming, and score $c_j\leftarrow \mathbf{W}_5\text{ReLU}(\mathbf{W}_4s_j^{(t)}) $ is generated for each answer candidate $j$. 

The input to the termination gate in step $t$ is the state representation of all possible answers, $s_j^{(t)}, j=1,2,...,r$. We do not use separate termination gates for each answer candidate. This is to restrain the size of the action space and variance in training. Since answers are mutable, the input weights for each answer candidate fed into the gate softmax are the same.

\begin{align*}
	p_t,1-p_t&\leftarrow \text{softmax}\left(\sum_{j=1}^r \mathbf{W_6}s_j^t\right)\\
	T_t&\sim \text{Bernoulli}(p_t).\\
\end{align*}

\noindent\textbf{Answer Prediction.} Finally, an answer prediction is drawn from the softmax distribution over the scores of each answer candidate:
\(C\sim\text{Softmax}\left(c_1,c_2,...,c_r\right). \)

\begin{table*}[ht]
	\centering
	\begin{tabular}{|l|ccc|ccc|ccc|}
		\hline
		Dataset & \multicolumn{3}{c|}{RACE-M} & \multicolumn{3}{c|}{RACE-H} & \multicolumn{3}{c|}{RACE} \\ \hline
		Subset                 & Train & Dev & Test & Train & Dev & Test & Train & Dev & Test    \\ \hline
		\# passages        & 6,409    & 368   & 362    & 18,728   & 1,021       & 1,045  &25,137 &1,389& 1,407  \\
		\# questions       & 25,421   & 1,436  & 1,436   & 62,445   & 3,451       & 3,498  &87,866 &4,887 &4,934  \\
		\hline
		Avg. Passage Len  & \multicolumn{3}{c|}{231.1}  & \multicolumn{3}{c|}{353.1}               & \multicolumn{3}{c|}{321.9}  \\
		Avg. Question Len & \multicolumn{3}{c|}{9.0}                 & \multicolumn{3}{c|}{10.4}               & \multicolumn{3}{c|}{10.0}    \\
		Avg. Option Len & \multicolumn{3}{c|}{3.9}                 & \multicolumn{3}{c|}{5.8}                 & \multicolumn{3}{c|}{5.3}    \\
		Vocab size        & \multicolumn{3}{c|}{32,811}              & \multicolumn{3}{c|}{125,120}               & \multicolumn{3}{c|}{136,629} \\
		\hline
	\end{tabular}
	\caption{Statistics on the RACE dataset from \cite{lai2017race}.}
	\label{tab:stat_race}
\end{table*}

\begin{table}[t!]
	\begin{center}
		\begin{tabular}{c|@{\hskip1pt}c|@{\hskip1pt} c|@{\hskip1pt}c}
			\hline \bf Model & RACE-M & RACE-H & RACE  \\ \hline
			Sliding Window & 37.3 & 30.4 & 32.2\\
			Stanford AR & 44.2 & 43.0 & 43.3\\
			GA & 43.7 & 44.2 & 44.1\\
			ElimiNet & N/A & N/A & 44.5 \\
			\textbf{DFN} & {\bf51.5} & \textbf{45.7} & \textbf{47.4} \\
			\hline
			GA$^*$ & N/A & N/A & 45.9\\
			ElimiNet$^*$ & N/A & N/A & 46.5 \\
			GA$^*$+ElimiNet$^*$ & N/A & N/A & 47.2 \\
			\textbf{DFN$^*$} & {\bf 55.6} & \textbf{49.4} & \textbf{51.2}\\
			\hline
		\end{tabular}
	\end{center}
	\caption{\label{tab:performance} Accuracy\% of DFN compared to baseline methods on RACE test sets. Results of the baseline models came from \cite{lai2017race} and unpublished ElimiNet
		($^*$ indicates ensemble models).
	}
	
\end{table}

\subsection{\label{sec:trainDetail} Training Details}
Since the strategy choice and termination steps are discrete random variables, DFN cannot be optimized by backpropagation directly. Instead, we see strategy choice $G$, termination decision $T_t$ and final prediction $C$ as policies, and use the REINFORCE algorithm \cite{williams1992simple} to train the network. 
Let $T$ be the actual skimming steps taken, i.e., $T=\min\{t:T_t=1\}$. We define the reward $r$ to be 1 if $C$ (final answer) is correct, and 0 otherwise. Each possible value pair of $(C,G,T)$ corresponds to a possible episode, which leads to $r\cdot n \cdot \mathcal{T}$ possible episodes. Let $\pi(c,g,t;\theta)$ be any policy parameterized by DFN parameter $\theta$, and $J(\theta)=E_{\pi}[r]$ be the expected reward. Then:

\begin{align}
	&\nabla_\theta J(\theta)\nonumber\\
	=&E_{\pi(g,c,t;\theta)}\left[\nabla_\theta\log \pi(c,g,t;\theta)(r-b)\right]\nonumber\\
	=&\sum_{g,c,t}\pi(g,c,t;\theta)\left[\nabla_\theta\log \pi(c,g,t;\theta)(r-b)\right].\label{eqn:gradient}
\end{align}
where $b$ is a critic value function. Following \cite{shen2017reasonet}, we set $b=\sum_{g,c,t}\pi(g,c,t;\theta)r$ and replace the $(r-b)$ term above by $(r/b-1)$ to achieve better performance and stability.

\section{\label{sec:expr}Experiments}


To evaluate the proposed DFN model, we conducted experiments on the RACE dataset. Statistics of the training/dev/test data are provided in Table \ref{tab:stat_race}. In this section, we present the experimental results, with a detailed analysis on the dynamic selection of strategies and multi-step reasoning. An ablation study is also provided to demonstrate the effectiveness of dynamic fusion and reasoning in DFN.

\subsection{Parameter Setup} 
Most of our parameter settings follow \cite{wang2017bilateral} and \cite{shen2017reasonet}. We use (\ref{eqn:gradient}) to update the model, and use ADAM \cite{kingma2014adam} with a learning rate of 0.001 and batch size of 64 for optimization. 
A small dropout rate of 0.1 is applied to each layer. For word embedding, we use 300-dimension GloVe \cite{pennington2014glove} embedding from the 840B Common Crawl corpus. The word embeddings are not updated during training. The character embedding has 20 dimensions and the character LSTM has 50 hidden units.  All other LSTMs have a hidden dimension of 100. The maximum reasoning step $\mathcal{T}$ is set to 5. We limit the length of passage/question/answer to a maximum of 500/100/100 for efficient computation. We also train an ensemble model of 9 DFNs using randomly initialized parameters. Training usually converges within 20 epochs. The model is implemented with Tensorflow \cite{abadi2016tensorflow} and the source code will be released upon paper acceptance.

\begin{table}[bt!]
	\begin{center}
		\begin{tabular}{|c|c|}
			\hline \bf Keyword & Dominant Strategy \& Steps  \\ \hline
			``\_'' & Integral Attention, Step 0  (93\%)\\
			``not''& Answer-only Attention, Step 5 (75\%)\\
			``except'' & Answer-only Attention, Step 5 (76\%)\\
			\hline
		\end{tabular}
	\end{center}
	\caption{\label{tab:keyword} Examples of attention and reasoning step choices associated with question types. ``\_'' is the placeholder for filling in answers (cf. Figure \ref{fig:gatechoice}).
	}
	
\end{table}

\begin{figure*}[ht!]
	\begin{center}
		\includegraphics[width=1.0\textwidth]{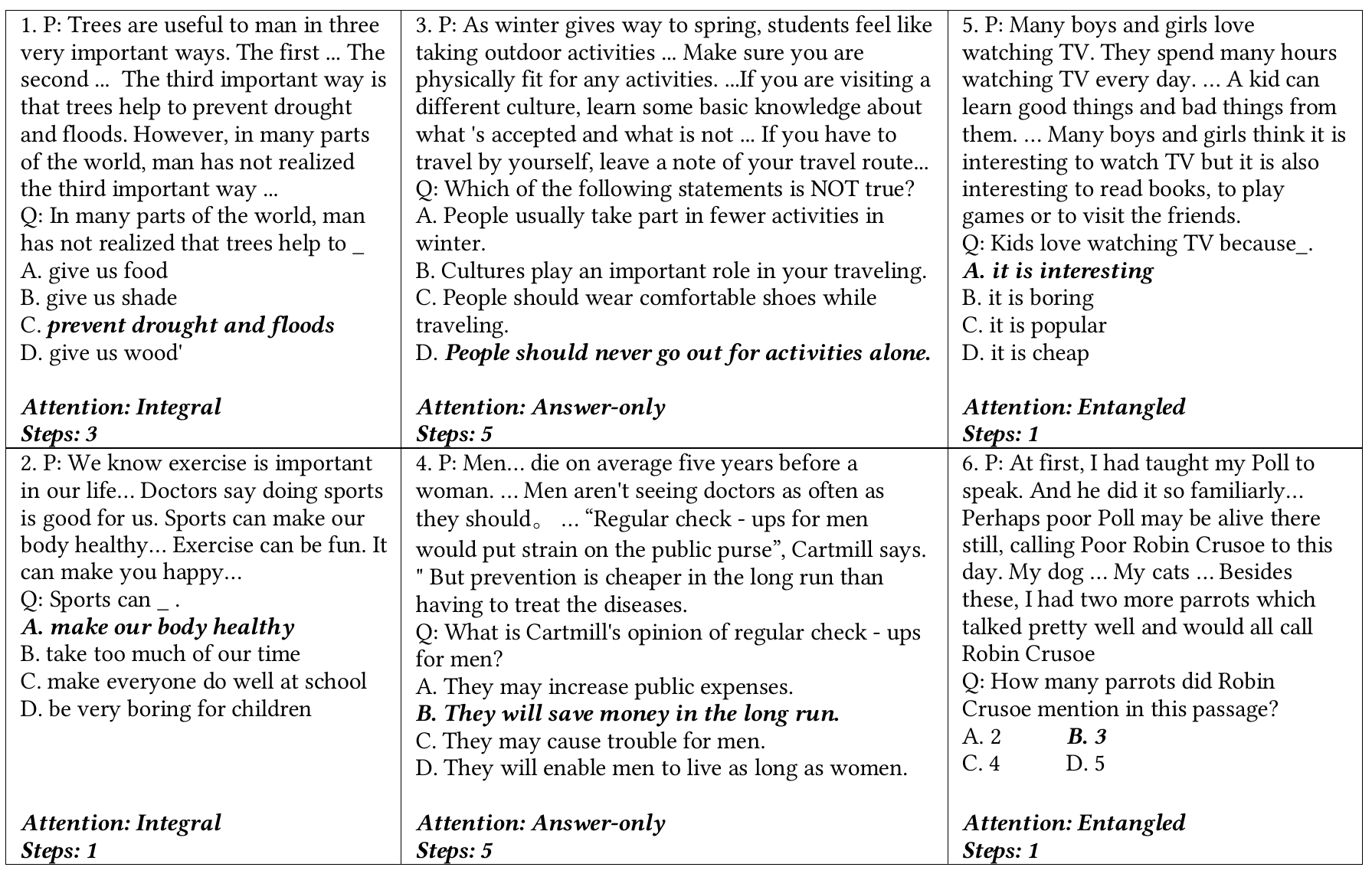}	
	\end{center}
	
	\caption{\label{fig:gatechoice}Examples of DFN's dynamic selection on attention strategy and reasoning steps. Correct answers are \textbf{\textit{bold and Italic}}.
	}

\end{figure*}

\subsection{Model Performance}


Table \ref{tab:performance} shows a comparison between DFN and a few previously proposed models. All models were trained with the full RACE dataset, and tested on RACE-M and RACE-H, respectively. As shown in the table, on RACE-M, DFN leads to a 7.8\% and 7.3\% performance boost over GA and Stanford AR, respectively. On RACE-H, the outperformance is 1.5\% and 2.7\%. The ensemble models also gained a performance boost of 4-5\% comparing to previous methods. We suspect that the lower gain on RACE-H might result from the higher level of difficulty in those questions in RACE-H, as well as ambiguity in the dataset. Human performance drops from 85.1 on RACE-M to 69.4 on RACE-H, which indicates RACE-H is very challenging even for human.

Figure \ref{fig:gatechoice} shows six randomly-selected questions from the dataset that DFN answered correctly, grouped by their attention strategies. Recall that the three attention strategies proposed for this task are: 1) Integral Attention for short answers; 2) Answer-only Attention for long answers; and 3) Entangled Attention for deeper reasoning. 
Question 1 and 2 in Figure \ref{fig:gatechoice} present two examples that used Integral Attention. In both of the questions, the question and answer candidates are partial sentences. So the system chose Integral Attention in this case. In the first question, DFN used 3 steps of reasoning, which indicates the question requires some level of reasoning (e.g., resolving coreference of ``the third way''). In the second question, the combined sentence comes directly from the passage, so DFN only used 1 step of reasoning.

Question 3 and 4 in Figure \ref{fig:gatechoice} provide two instances that use answer-only attentions. As shown in these examples, Answer-only attention usually deals with long and natural language answer candidates. Such answers cannot be derived without the model reading through multiple sentences in the passage, and this requires multi-step reasoning. So in both examples, the system went through 5 steps of reasoning. 

Question 5 and 6 in Figure \ref{fig:gatechoice} show two examples that used the Entangled Attention. Both questions require a certain level of reasoning. Question 5 asks for the causes of a scenario, which is not explicitly mentioned in the passage. And question 6 asks for a counting of concepts, which is implicit and has to be derived from the text as well. For both cases, the entangled attention was selected by the model. As for the reasoning steps, we find that for the majority of questions that use Entangled Attention, DFN only uses one reasoning step. This is probably because entangled attention is powerful enough to derive the answer.

We also examined the strategy choices with respect to certain keywords. For each word $w$ in vocabulary, we computed the distribution $\Pr[G,T|w\in Q]$, 
i.e., the conditional distribution of strategy and step when $w$ appeared in the question. Table \ref{tab:keyword} provides some keywords and their associated dominant strategies and step choices.
The results validate the assumption that DFN dynamically selects specific attention strategy based on different question types. For example, the underline ``\_'' indicates that the question and choice should be concatenated to form a sentence. This led to Integral Attention being most favorable when ``\_'' is present. In another example, ``not'' and ``except'' usually appear in questions like ``Which of the following is not TRUE''.
Such questions usually have long answer candidates that require more reasoning. So Answer-only Attention with Reasoning Step\#5 became dominant. 

\begin{table}[b!]
	\begin{center}
		\begin{tabular}{|l|c|c|}
			\hline \bf Model & Acc (dev)\% & Acc (test)  \\ \hline
			i) DFN & 52.9 & 50.6\\
			ii) W/o DF & 52.2& 49.5\\
			iii) W/o MR & 52.5& 49.4\\
			iv) W/o DF, MR & 52.4& 49.0*\\
			\hline
		\end{tabular}
	\end{center}
	\caption{\label{tab:ablation} Ablation studies of DFN for Dynamic Fusion (DF) and multi-step reasoning (MR). Results are from 3 ensembles to avoid variance in training. * indicates that the improvement of DFN over the ablation model is significant at the level of p<0.01.} 
	
\end{table}



\subsection{Ablation Studies} 
For ablation studies, we conducted experiments with 4 different model configurations:
\begin{enumerate}[i)]
	\item The full DFN model with all the components aforementioned.
	\item \label{point:noDF}DFN without dynamic fusion (DF). We dropped the Strategy Gate $G$, and used only one attention strategy in the Dynamic Fusion Layer. 
	\item DFN without multi-step reasoning (MR). Here we dropped the Answer Scoring Module, and used the output of Dynamic Fusion Layer to generate a score for each answer. 
	\item \label{point:noDFMR}DFN without DF and MR. 
\end{enumerate}

To select the best strategy for each configuration, we trained 3 different models for ii) and iv), and chose the best model based on their performance on the dev set. This explains the smaller performance gap between the full model and ablation models on the dev set than that on the test set. Experimental results show that for both ii) and iv), the Answer-Only Attention gave the best performance. 

To avoid variance in training and provide a fair comparison, 3 ensembles of each model were trained and evaluated on both dev and test sets. As shown in Table \ref{tab:ablation}, the DFN model has a ~1.6\% performance gain over the basic model (without DF and MR). This performance boost was contributed by both multi-step reasoning and dynamic fusion. When omitting DF or MR alone, the performance of DFN model dropped by 1.1\% and 1.2\%, respectively. 

To validate the effectiveness  of the DFN model, we also performed a significance test and compared the full model with each ablation model. The null hypothesis is: the full DFN model has the same performance as the ablation model. As shown in Table \ref{tab:ablation}, the combination of DF and MR leads to an improvement with a statistically significant margin in our experiments, although neither DF or MR can, individually.

\section{Conclusion}

In this work, we propose a novel neural model - Dynamic Fusion Network (DFN), for MRC. For a given input sample, DFN can dynamically construct an model instance with a sample-specific network structure by picking an optimal attention strategy and an optimal number of reasoning steps on the fly. The capability allows DFN to adapt effectively to handling questions of different types.
By training the policy of model construction with reinforcement learning, our DFN model can substantially outperform previous state-of-the-art MRC models on the challenging RACE dataset. Experiments show that by marrying dynamic fusion (DF) with multi-step reasoning (MR), the performance boost of DFN over baseline models is statistically significant. For future directions, we plan to incorporate more comprehensive attention strategies into the DFN model, and to apply the model to other challenging MRC tasks with more complex questions that need DF and MR jointly. Future extension also includes constructing a ``composable'' structure on the fly - by making the Dynamic Fusion Layer more flexible than it is now. 

\bibliography{yichongref}
\bibliographystyle{acl_natbib}

\end{document}